\documentclass[conference]{IEEEtran}
\IEEEoverridecommandlockouts
\usepackage{cite}
\usepackage{amsmath,amssymb,amsfonts}
\usepackage{algorithmic}
\usepackage{graphicx}
\usepackage{textcomp}
\usepackage{comment}
\usepackage{xcolor}
\usepackage{wrapfig}
\usepackage{subcaption}
\usepackage{multirow}
\usepackage{adjustbox}
\usepackage{babel,blindtext}

\def\BibTeX{{\rm B\kern-.05em{\sc i\kern-.025em b}\kern-.08em
    T\kern-.1667em\lower.7ex\hbox{E}\kern-.125emX}}
\begin{document}

\title{Investigating Fairness of Ocular Biometrics Among Young, Middle-Aged, and Older Adults
}

\author{\IEEEauthorblockN{Anoop Krishnan, Ali Almadan and Ajita Rattani}
\IEEEauthorblockA{\textit{School of Computing} \\
\textit{Wichita State University, USA}\\
\{axupendrannair, aaalmadan\}@shockers.wichita.edu; ajita.rattani@wichita.edu}
}

\maketitle
\begin{abstract}
A number of studies suggest bias of the face biometrics, i.e., face recognition and soft-biometric estimation methods, across gender, race, and age-groups. There is a recent urge to investigate the bias of different biometric modalities toward the deployment of fair and trustworthy biometric solutions. Ocular biometrics has obtained increased attention from academia and industry due to its high accuracy, security, privacy, and ease of use in mobile devices. A recent study in $2020$ also suggested the fairness of ocular-based user recognition across males and females.
This paper aims to evaluate the fairness of ocular biometrics in the visible spectrum among age-groups; young, middle, and older adults. Thanks to the availability of the latest large-scale $2020$ UFPR ocular biometric dataset, with subjects acquired in the age range $18$ - $79$ years, to facilitate this study. 
Experimental results suggest the overall equivalent performance of ocular biometrics across gender and age-groups in user verification and gender-classification. Performance difference for older adults at lower false match rate and young adults was noted at user verification and age-classification, respectively. This could be attributed to inherent characteristics of the biometric data from these age-groups impacting specific applications, which suggest a need for advancement in sensor technology and software solutions. 


  
\end{abstract}


\section{Introduction}

Decades of research have been conducted in extracting representative features from biometric modalities, such as the face, fingerprint, and ocular region, for user recognition and soft-biometric estimation such as gender, race, and age-group~\cite{jain04,RAJA2020103979,alonsofernandez2021facial}. Biometric technology has been widely adopted in forensics, surveillance, border-control, human-computer interaction, anonymous customized advertisement system, and image retrieval systems. 
\begin{figure}[t]{}
\centering
\includegraphics[width=1\linewidth]{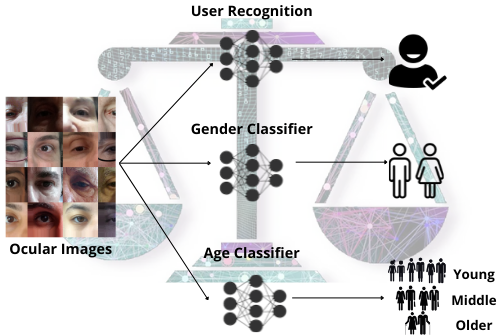}
\caption{\label{fig:schema} \small{Illustration of this Study on Fairness of the Ocular Biometrics Among Young, Middle and Older Adults}.}
\end{figure} 

However, over the last few years, \emph{fairness} of the automated face-based recognition and soft-biometric prediction methods have been questioned across gender, race, and age-groups~\cite{9001031,9086771,Buolamwini18, DBLP:conf/icmla/KrishnanAR20}. Fairness is defined as the absence of any prejudice or favoritism towards a group based on their inherent or acquired characteristics. Specifically, the majority of these studies raise the concern of higher error rates of face-based recognition and soft-biometric estimation methods for darker-skinned people, women, and older adults. This has led to the temporary ban of face recognition technology across various cities-first San Francisco and then others.  

For the \emph{deployment of fair and trustworthy biometric technology}, \textbf{two lines of research} are being pursued: (a) thorough \emph{investigation of different biometric modalities} for bias analysis across demographic variables~\cite{9001031, 9086771, krishnan2020probing, DBLP:conf/aaai/0001ASNV20,9287219}, and (b) development of \emph{fairness-aware} biometric systems~\cite{DBLP:journals/corr/abs-1902-00334,Jain2021,10.1007/978-3-030-58526-6_20}. 

For person authentication and soft-biometric prediction, \emph{ocular biometrics in the visible spectrum} include scanning regions in and around the eye, such as the iris, conjunctival and episcleral vasculature, and periocular region~\cite{RATTANI20171, VISOB,RAJA2020103979,9211002,alonsofernandez2021facial} has been well-established.  Due to its high accuracy, privacy, the convenience of capture with a standard RGB camera in mobile devices, and in the presence of a facial mask~\cite{shreyas}, this modality has received a lot of attention from academia and industry. 
A number of deep learning methods based on fine-tuned convolutional neural networks (CNNs) have been proposed for \emph{user authentication as well as soft-biometric estimation from ocular regions}~\cite{RATTANI20171,VISOB,RAJA2020103979,9211002,alonsofernandez2021facial}. Studies~\cite{9211002,alonsofernandez2021facial} suggest gender classification accuracy from the ocular region in the range [71\%, 90\%] using fine-tuned VGG-16, ResNet-50, DenseNet, and MobileNet-V2.  Studies on age-group classification from ocular images~\cite{9211002,alonsofernandez2021facial,8272766}, suggest exact and 1-off accuracy values in the range [36.4\%, 46.97\%]  and [48.4\%, 80.96\%], respectively. These results from existing studies suggest that equivalent performance (with 2\% to 5\% accuracy difference) could be obtained in gender and age classification from the ocular region over face biometrics~\cite{alonsofernandez2021facial}. Datasets such as MICHE-I~\cite{10.1016/j.patrec.2015.02.009}, VISOB~\cite{VISOB} and lately UFPR $2020$~\cite{zanlorensi2020ufprperiocular} have been assembled for research and development in RGB ocular-based user authentication and attribute classification. VISOB and UFPR are the latest large-scale RGB ocular biometrics datasets consisting of $550$ and $1122$ subjects, respectively. 

However, to date, the \emph{fairness of ocular biometrics} has not been well studied. The study by Krishnan et al.~\cite{krishnan2020probing} in $2020$ is the first study that investigated the fairness of ocular-based user recognition and gender classification across males and females on the VISOB RGB ocular biometric dataset. Reported results on \emph{subject-disjoint gender-balanced} dataset suggest that, in contrary to the existing studies on face recognition, equivalent authentication performance could be obtained for males and females based on the ocular region. 

The aim of this study is to \emph{investigate the fairness of the ocular biometrics in visible spectrum} among age-groups; young, middle-aged, and older adults. Following the standard guidelines~\cite{10.1093/geront/42.1.92}, a young person is in the age range $18$ to $39$, middle-aged in $40$ to $59$, and an older adult is in the age range $60$ to $79$ years. This study is important towards understanding the fairness of ocular technology among age-groups. The \emph{challenge} include most of the existing publicly available ocular biometric datasets in the visible spectrum, such as VISOB~\cite{VISOB}, acquires only young population. Further, age and gender information is not publicly available. Thanks to the availability of the latest $2020$ UFPR ocular biometric dataset~\cite{zanlorensi2020ufprperiocular}, consisting of $1122$ subjects in the age range $18-79$ annotated with gender and age information, which facilitates this study.
Figure~\ref{fig:schema} illustrates our study on evaluating the fairness of the ocular biometrics (user verification, gender- and age-classification) among age-groups.

The contributions of this paper are as follows:
\begin{itemize}
\item Evaluation of the \textbf{fairness} of ocular-based user authentication, gender- and age-group classification algorithms among young, middle-aged, and older adults using fine-tuned ResNet-50,  MobileNet-V2, ShuffleNet-V2, and EfficientNet-B0 models. 

\item The use of \textbf{Explainable AI (XAI)} based Gradient-weighted Class Activation Mapping (Grad-CAM) visualization~\cite{8237336} to understand the distinctive image regions used by the CNN models in classifying different age-groups.
\end{itemize}
This paper is organized as follows: Sections II discusses the dataset and implementation details. Fairness analysis of ocular-based user verification and soft-biometrics prediction is discussed in section III. Key findings are listed in section IV. Conclusions are drawn in section V.

\section{Dataset and Implementation Details}
In this section, we discuss the dataset used and the implementation details of this study.

\par
\begin{figure}[]
  \centering
  \subfloat[]{\includegraphics[width=0.12\textwidth]{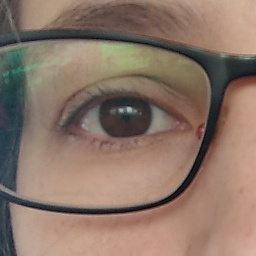}\label{lfy}}\hspace{0.1cm}
    \subfloat[]{\includegraphics[width=0.12\textwidth]{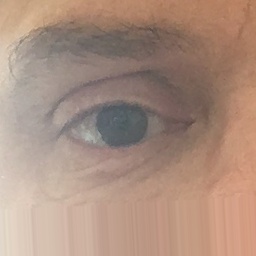}\label{lmo}}\hspace{0.1cm}
     \subfloat[]{\includegraphics[width=0.12\textwidth]{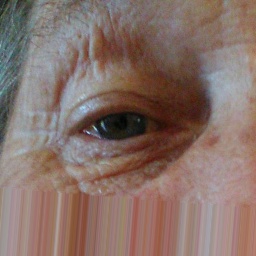}\label{lfo}}\hspace{0.1cm}
     \subfloat[]{\includegraphics[width=0.12\textwidth]{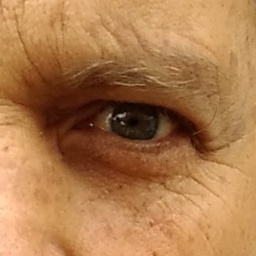}\label{rmo}}\hspace{0.1cm}
    \subfloat[]{\includegraphics[width=0.12\textwidth]{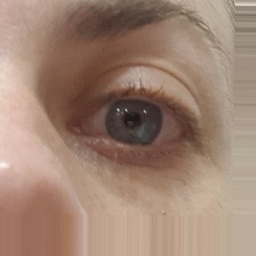}\label{rfm}}\hspace{0.1cm}
    \subfloat[]{\includegraphics[width=0.12\textwidth]{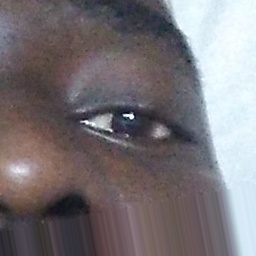}\label{rmy}}
  \caption{Sample of left and right ocular images from the UFPR dataset~\cite{zanlorensi2020ufprperiocular} acquired across gender, race and age.}  
  \label{sample}
\end{figure}

 \textbf{UFPR-Periocular~\cite{zanlorensi2020ufprperiocular}:} 
This is the most recent $2020$ RGB ocular biometric dataset, which includes $33,660$ ocular samples from $1,122$ subjects captured by $196$ different mobile devices. The dataset is collected from subjects across race, age, and gender. 
The ocular images  were  normalized  in  terms  of  rotation  and  scale  using  the manual annotations of the corners of the eyes by the authors. Sample ocular images are shown in Figure~\ref{sample}. Authors~\cite{zanlorensi2020ufprperiocular} have reported an \textit{average accuracy of $97.80$\% and $84.34$\% for gender and age-group classification, respectively, on this dataset}. This high accuracy is attributed to the normalizing step.

For the purpose of our study, all images were resized to $224 \times 224$. Using the standard guidelines~\cite{10.1093/geront/42.1.92}, the complete dataset is classified into three age-groups: namely young ($18$ to $39$), middle-aged ($40$ to $59$), and older adults ($60$ to $79$), using the publicly available age information. 
   All the experiments were conducted using \textit{subject-disjoint gender and age-group balanced} training and testing sets. Subject-disjoint means that the subjects do not overlap between the training and testing set. Gender and age-group balanced training and testing sets were obtained by discarding samples from young adults and using random data augmentation across age-groups. The dataset could not be race-balanced due to the unavailability of race information.

For user authentication and gender classification, following the protocol used in UFPR dataset~\cite{zanlorensi2020ufprperiocular}, we used the fine-tuned version of widely used ResNet-50, mobile-friendly  MobileNet-V$2$, ShuffleNet-V$2$, and EfficientNet-B$0$ models. The choice of these models is motivated by the analysis of the bias across different CNN architectures ranging from widely popular heavy-weight (ResNet) to light-weight mobile friendly models (MobileNet-V$2$, ShuffleNet-V$2$, and EfficientNet-B$0$).
These CNN models are fine-tuned on randomly selected gender and age-group balanced subset from $780$ participants. 
Subject-disjoint, gender and age-group-balanced subset selected from $342$ subjects are used as the test set for authentication and $260$ subjects ($130$ males and females) for gender classification. After the last convolutional layer, Batch Normalization (BN), dropout, and the fully connected layer of size $512$, and the final output layer were added for user recognition. Deep features of size $512$ were extracted from the five template and test sets per subject and matched using cosine similarity for the computation of matching scores. \textbf{Note this is not a template aging study so template and test image belong to the same age-group}. 
The fully connected layers of size 1024, 512, and 512, as well as dropout layers, were included after the last convolutional layer for the gender classification, followed by the final binary classification layer. Age-group classification models (ResNet-50, mobile-friendly  MobileNet-V$2$, ShuffleNet-V$2$, and EfficientNet-B$0$) were fine-tuned on balanced subset selected from $432$ subjects and evaluated on subset from $132$ subjects, across $6$ age-groups, namely $18-29$, $30-39$, $40-49$, $50-59$, $60-69$ and $70-79$. All models were trained using an early stopping mechanism using Adagrad optimizer, Glorot initialization and the cross-entropy loss function.

\section{Fairness of Ocular Biometrics among Age-groups}
\par The fairness of ocular-based CNN models trained for user recognition, gender and age-group classification among young, middle-aged, and older adults is analyzed in this section. Equal Error Rate (EER), False Non-Match Rate (FNMR) at lower False Match Rate (FMR) points for user recognition, and classification accuracy for soft-biometrics are reported following the standard biometric evaluation procedure.
\vspace{0.1 cm}


\begin{table*}[t]
 \caption{EER and FNMR at $0.01$ and $0.1$ FMR for user authentication using CNN models for Left (L), Right (R) ocular region, and their score-level fusion (L+R) for Young, Middle-Aged and Older adults evaluated on balanced version of UFPR~\cite{zanlorensi2020ufprperiocular} ocular datasets.}
\centering
\resizebox{0.9\linewidth}{!}{%
\begin{tabular}{|c|c|ccc|ccc|ccc|}
\hline
\textbf{CNN}              & \textbf{Age-group} & \multicolumn{3}{c|}{\textbf{EER(\%)}}                                        & \multicolumn{6}{c|}{\textbf{FNMR(\%) @ FMR}}                                                                                                         \\ \hline
                          &                    &                            &                            &                & \multicolumn{3}{c|}{\textbf{0.01}}                                       & \multicolumn{3}{c|}{\textbf{0.1}}                                        \\ \cline{3-11} 
                          &                    & \textbf{L}                 & \textbf{R}                 & \textbf{L + R} & \textbf{L}                 & \textbf{R}                 & \textbf{L + R} & \textbf{L}                 & \textbf{R}                 & \textbf{L + R} \\ \hline
\textbf{ResNet-50}        & \textbf{Young}     & \multicolumn{1}{c|}{8.60}  & \multicolumn{1}{c|}{9.52}  & 9.06           & \multicolumn{1}{c|}{37.63} & \multicolumn{1}{c|}{35.35} & 36.49          & \multicolumn{1}{c|}{53.74} & \multicolumn{1}{c|}{54.03} & 53.89          \\ \cline{2-11} 
                          & \textbf{Middle-Aged}    & \multicolumn{1}{c|}{8.62}  & \multicolumn{1}{c|}{9.08}  & 8.85           & \multicolumn{1}{c|}{30.76} & \multicolumn{1}{c|}{25.04} & 27.9           & \multicolumn{1}{c|}{52.23} & \multicolumn{1}{c|}{48.55} & 50.39          \\ \cline{2-11} 
                          & \textbf{Older}     & \multicolumn{1}{c|}{11.01} & \multicolumn{1}{c|}{11.00} & 11.01          & \multicolumn{1}{c|}{15.47} & \multicolumn{1}{c|}{19.34} & 17.405         & \multicolumn{1}{c|}{30.67} & \multicolumn{1}{c|}{30.67} & 30.67          \\ \hline
\textbf{MobileNet-V2}     & \textbf{Young}     & \multicolumn{1}{c|}{7.75}  & \multicolumn{1}{c|}{7.32}  & 7.54         & \multicolumn{1}{c|}{33.21} & \multicolumn{1}{c|}{26.11} & 29.66          & \multicolumn{1}{c|}{51.61} & \multicolumn{1}{c|}{48.28} & 49.95          \\ \hline
                          & \textbf{Middle-Aged}    & \multicolumn{1}{c|}{9.08}  & \multicolumn{1}{c|}{8.68}  & 8.88           & \multicolumn{1}{c|}{29.18} & \multicolumn{1}{c|}{28.14} & 28.66          & \multicolumn{1}{c|}{51.17} & \multicolumn{1}{c|}{51.31} & 51.24          \\ \cline{2-11} 
                          & \textbf{Older}     & \multicolumn{1}{c|}{8.04}  & \multicolumn{1}{c|}{9.63}  & 8.84           & \multicolumn{1}{c|}{18.54} & \multicolumn{1}{c|}{13.47} & 16.005         & \multicolumn{1}{c|}{39.47} & \multicolumn{1}{c|}{36.00} & 37.74          \\ \hline
\textbf{ShuffleNet-V2} & \textbf{Young}     & \multicolumn{1}{c|}{\textbf{6.93}}  & \multicolumn{1}{c|}{6.96}  & 6.95           & \multicolumn{1}{c|}{37.63} & \multicolumn{1}{c|}{38.09} & 37.86          & \multicolumn{1}{c|}{56.44} & \multicolumn{1}{c|}{56.26} & 56.35          \\ \hline
                          & \textbf{Middle-Aged}       & \multicolumn{1}{c|}{8.32}  & \multicolumn{1}{c|}{9.25}  & 8.79           & \multicolumn{1}{c|}{37.38} & \multicolumn{1}{c|}{46.07} & \textbf{41.73}         & \multicolumn{1}{c|}{55.08} & \multicolumn{1}{c|}{61.10} & \textbf{58.09}          \\ \cline{2-11} 
                          & \textbf{Older}     & \multicolumn{1}{c|}{9.72}  & \multicolumn{1}{c|}{8.18}  & 8.95           & \multicolumn{1}{c|}{30.53} & \multicolumn{1}{c|}{30.80} & 30.67       & \multicolumn{1}{c|}{44.93} & \multicolumn{1}{c|}{53.47} & 49.20           \\ \hline
\textbf{EfficientNet-B0}          & \textbf{Young}     & \multicolumn{1}{c|}{7.64}  & \multicolumn{1}{c|}{9.61}  & 8.63           & \multicolumn{1}{c|}{32.54} & \multicolumn{1}{c|}{27.00} & 29.77          & \multicolumn{1}{c|}{55.40} & \multicolumn{1}{c|}{48.38} & 51.89          \\ \hline
                          & \textbf{Middle-Aged}       & \multicolumn{1}{c|}{6.95}  & \multicolumn{1}{c|}{9.03}  & 7.99           & \multicolumn{1}{c|}{38.07} & \multicolumn{1}{c|}{26.69} & 32.38          & \multicolumn{1}{c|}{49.12} & \multicolumn{1}{c|}{49.31} & 49.22          \\ \cline{2-11} 
                          & \textbf{Older}     & \multicolumn{1}{c|}{9.34}  & \multicolumn{1}{c|}{12.08} & 10.71          & \multicolumn{1}{c|}{20.80} & \multicolumn{1}{c|}{23.33} & 22.065         & \multicolumn{1}{c|}{39.73} & \multicolumn{1}{c|}{41.74} & 40.74          \\ \hline
\end{tabular}%
}
\label{tab:recon_res}
\end{table*}

\subsection{Fairness among Age-groups in User Verification}
Table~\ref{tab:recon_res} shows the subject-disjoint user verification (recognition) performance of the CNN models evaluated using False Non-Match Rates (FNMRs) at $0.01$ and $0.1$ False Match Rates (FMR), and Equal Error Rate (EER) among Young, Middle, and Older adults. 

As can be seen, ShuffleNet-V2 and MobileNet-V2 obtained the lowest average EER of about $8.32$\% for the three age-groups. ResNet-50 and EfficientNet-B0 obtained a similar average EER of about $9.43$\%. 
In terms of FNMR, it can be seen that ShuffleNet-V2 obtained the highest FNMR of $41.73$\% and $58.09$\% at FMRs of 0.01 and 0.1, respectively. The FNMR of the other four models (ResNet-50, MobileNet-V2, and EfficientNet-B0) were in the range of [$16.01$-$36.49$]\% at FMR of 0.01 and an average of $48.29$\% was obtained at FMR of $0.1$. 
\par  The mean EER of $8.04$\% was obtained for Young adults. Middle-aged and Older adults obtained a mean EER of $9.01$\% and $9.88$\%, respectively. Therefore, Young adults outperformed Middle-Aged and Older adults in user recognition by a slight difference of $1\%$ EER.
Across CNN models, the least EER of $6.95$\% was obtained for Young adults by ShuffleNet-V2.
The maximum EER of $11.01$\% was obtained for Older adults using the ResNet-50 model.

For FNMR@FMR=0.01, the Young and Middle-aged groups obtained identical mean performance of $33.0\%$, with the Older group FNMR dropping to $11.50\%$. Middle-aged adults, on the other hand, obtained a maximum FNMR@FMR=0.01 of $41.73\%$, which was lower than the Young group by $4\%$ and higher than the Older adults by $11.0\%$ (maximum value was obtained by ShuffleNet-V2 across age-groups). For FNMR@FMR=0.1, the same trend was observed, with Young and Middle-aged adults obtaining the maximum performance with an average of $52.6\%$. Their FNMR is $13.0$\% higher than the FNMR obtained by Older adults. An insignificant performance difference was noted across males and females which is also supported by our initial study in~\cite{krishnan2020probing}.

\emph{Overall, the Young and Middle-aged groups performed similarly across EER, FNMR@FMR=0.01, and FNMR@FMR=0.1, with the Young group showing a slight improvement. Older adults' overall performance dropped only by 1\% EER, but their FNMR dropped significantly at lower FMR points.} No consistency in performance is observed across the left and right ocular region, and their score level fusion (using the averaging rule) did not improve the performance over individual units.

\subsection{Fairness among Age-groups in Soft-biometrics}

\begin{table*}[]
\caption{\label{tab:left-gender}Accuracy of CNN-based Gender Classification on Left Ocular Region among Young (18 to 39 years), Middle (40 to 59 years) and Older Adults (60 to 79 years).}
\centering
\resizebox{0.9\linewidth}{!}{%
\begin{tabular}{|c|c|c|c|c|c|c|}
\hline
\multicolumn{1}{|c|}{\textbf{CNN}} & \multicolumn{2}{c|}{\textbf{Young}}                    & \multicolumn{2}{c|}{\textbf{Middle-Aged}}             & \multicolumn{2}{c|}{\textbf{Older}}                    \\ \hline
\multicolumn{1}{|l|}{} & \textbf{Male [\%]}                   & \textbf{Female [\%]}                     & \textbf{Male [\%]}                     & \textbf{Female [\%]}                     & \textbf{Male [\%]}                     & \textbf{Female [\%]}                     \\ \hline
\textbf{ResNet-50}              & 98.39                & 98.19                & \textbf{100}                  & 96.67                 & 99.17             & 98.06                 \\ \hline
\textbf{MobileNet-V2}             & \textbf{99.97}         & \textbf{99.9}                & \textbf{100}                & \textbf{99.7}              & \textbf{100}                & \textbf{100}                 \\ \hline
\textbf{ShuffleNet-V2-50}             & 98.23          & 97.57             & 98.28                & 97.56                & 94.76                 & 98.89                \\ \hline
\textbf{EfficientNet-B0}        &  95.89           & 97.54             & 98.58              & 96.44                & 95.23                & 86.94 \\ \hline
\end{tabular}%
}
\end{table*}

\begin{table*}[t]
\caption{\label{tab:right-gender}Accuracy of the CNN-based Gender Classification on Right Ocular Region among Young (18 to 39 years), Middle (40 to 59 years) and Older Adults (60 to 79 years).}
\centering
\resizebox{0.9\linewidth}{!}{%
\begin{tabular}{|c|c|c|c|c|c|c|}
\hline
\multicolumn{1}{|c|}{\textbf{CNN}} & \multicolumn{2}{c|}{\textbf{Young}}                    & \multicolumn{2}{c|}{\textbf{Middle-Aged} }  & \multicolumn{2}{c|}{\textbf{Older}}                    \\ \hline
\multicolumn{1}{|l|}{} & \textbf{Male [\%]}                  & \textbf{Female [\%]}                     & \textbf{Male [\%]}                     & \textbf{Female [\%]}                     & \textbf{Male [\%]}                     & \textbf{Female [\%]}                     \\ \hline
\textbf{ResNet-50}            & 97.98         & \textbf{99.61}         & 97.07           & \textbf{99.78}      & 92.86        & \textbf{98.61}         \\ \hline
\textbf{MobileNet-V2}          & 96.84        & 92.35          & 95.86         & 98.66        & 94.76         & 97.5          \\ \hline
\textbf{ShuffleNet-V2-50}       & \textbf{98.51}           & 98.91        & \textbf{98.79}      & 99.33        & \textbf{97.38}        & \textbf{98.61}          \\ \hline
\textbf{EfficientNet-B0}        &97.18           & 95.14              & 95.15               & 97.33               & 94.52                 & 90  \\ \hline
\end{tabular}
}
\end{table*}
\noindent \textbf{Gender Classification}: The accuracy of the CNN-based gender classification among age-groups are shown in Tables~\ref{tab:left-gender} and \ref{tab:right-gender}. From Tables~\ref{tab:left-gender} and \ref{tab:right-gender}, middle-aged adults outperformed the Young and Older adults by $1$\% and $1.4$\% for left and right ocular region. On an average, Females obtained an accuracy of $98.3$\% ($96.5$\%), $97.59$\% ($98.78$\%), and $95.97$\% ($96.18$\%) for Young, Middle-aged and Older Adults, respectively, for left (right) ocular region. On an average, Males obtained an accuracy of $98.12$\% ($97.63$\%), $99.2$\% ($96.71$\%), and $97.3$\% ($94.88$\%) for Young, Middle-aged and Older Adults, respectively, for left (right) ocular region. \textit{Males and Females performed equally with an insignificant accuracy gap of $0.09\%$}. From the Table~\ref{tab:left-gender}, maximum and minimum accuracy values for left ocular region are $99.9$\% and $86.94$\%, and that of right ocular region are $99.78$\% and $90$\% from Table~\ref{tab:right-gender}. Therefore, right ocular region outperformed left ocular region. Fusion of left and right ocular regions did not improve the classification accuracy. For the sake of space, fusion results are not reported. The obtained high gender classification accuracy rates are in the range reported by the authors on UFPR dataset~\cite{zanlorensi2020ufprperiocular}.

\begin{figure}[t]
  \centering
  \subfloat[]{\includegraphics[width=0.15\textwidth]{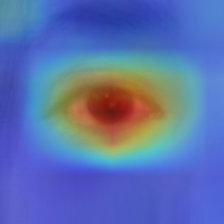}\label{tg_yf}}\hspace{0.1cm}
    \subfloat[]{\includegraphics[width=0.15\textwidth]{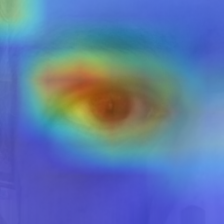}\label{tg_mm}}\hspace{0.1cm}
    \subfloat[]{\includegraphics[width=0.15\textwidth]{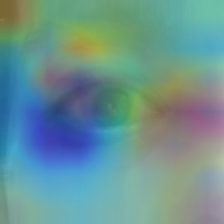}\label{tg_om}}
\caption{Averaged Grad-CAM~\cite{8237336} visualization of true and false gender classification for ResNet-$50$ based model. True Classification - (a) Young Female, and (b) Middle Aged Male. False Classification - (c)Young Female.} 
\label{GRAD-gender}
\end{figure}

Further, in order to understand the distinctive image regions used by CNN models in predicting gender across age-groups, we used \emph{Gradient-weighted Class Activation Mapping} (GRAD-CAM)~\cite{8237336} as a tool for XAI. GRAD-CAM uses the gradients of any target concept to generate a coarse localization map that highlights distinctive image regions used for making a decision/prediction. 
Figure~\ref{GRAD-gender} shows the GRAD-CAM visualization of the ResNet-50-based gender classifier. The highly activated region is shown by the red zone on the map, followed by green and blue zones. To get further insight, we averaged randomly selected heat maps of GRAD-CAM visualization for each gender. 
It can be seen that for females, \textbf{the pupil and sclera regions} are the highly activated regions used for gender classification, while for males, \textbf{the upper eyelid area} is used. The activated regions for males and females \emph{remain consistent} across the young, middle, and older adults for gender classification. The use of the non-ocular area in learning differentiating traits for gender classification is the most common cause of inaccurate classification (see Figure~\ref{GRAD-gender} (c)). The same observation is noted for other CNN models used in the study. In~\cite{kuehlkamp2018predicting}, Kuehlkamp et al. suggest that discriminative power of the iris texture for gender is weak and that the gender-related information is primarily in the periocular region which is supported by our observations as well.

\emph{Overall, equivalent performance was obtained for gender classification among age-groups. Middle-Aged adults slightly outperformed Young and Older adults by about a 1\% increment in gender classification from the ocular regions. GRAD-CAM visualization suggested gender-specific image region selection by CNN models for gender classification from ocular images}.

\noindent \textbf{Age-classification}: Table \ref{tab:young}, \ref{tab:middle} and \ref{tab:older} show the exact and $1$-off age-group classification accuracy of the fine-tuned ResNet-$50$, MobileNet-V$2$, ShuffleNet-V$2$ and EfficientNet-B$0$ among young, middle-aged and older adults on left and right ocular region. $1$-off means the predicted label is within the neighboring group of the true class. On an average, young adults have obtained an average exact ($1$-off) accuracy of $46.37$\% ($83.14\%$) with minimum and maximum of $29.9$\% ($57.9$\%) and $56.95$\% ($93.34$\%) for EfficientNet-B$0$ and MobileNet-V$2$. 
On an average, Middle-aged adults have obtained an average exact (1-off) accuracy of $27.5$\% ($69.45$\%) with minimum and maximum of $50.24$\% ($56.8$\%) and $33.32$\% ($80.08$\%) for EfficientNet-B$0$ and ResNet-$50$. 
On an average, Older adults have obtained an average exact (1-off) accuracy of $24.1$\% ($64.5$\%) with minimum and maximum of $4.9$\% ($29.65$\%) and $35.5$\% ($78.23$\%) for EfficientNet-B$0$ and MobileNet-V$2$. 
\par
\begin{figure}[t]
  \centering
  \subfloat[]{\includegraphics[width=0.15\textwidth]{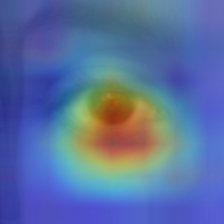}\label{fig:18-29_age}}\hspace{0.1cm}
  \subfloat[]{\includegraphics[width=0.15\textwidth]{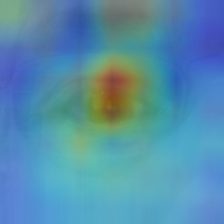}\label{fig:50-59_age}}
    \hspace{0.1cm}
    \subfloat[]{\includegraphics[width=0.15\textwidth]{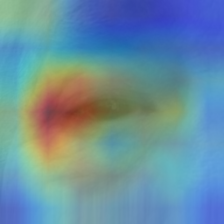}\label{fig:70-79_age}}
 \caption{Averaged Grad-CAM~\cite{8237336} visualization of true classification for ResNet-$50$ based Age-group Classification. (a) Young, (b) Middle-Aged, and (c) Older Adult.} 
 \label{GRAD_age}
\end{figure}

 Thus, young adults performed the best by $24\%$ and middle and older adults performed equivalently for both the ocular regions. Mostly, the left ocular region obtained better accuracy than the right ocular region. The fusion of the left and right ocular regions did not improve the accuracy rates. In most cases, Exact and $1$-off accuracy values differ by more than $40$\%. Further, Figure~\ref{GRAD_age} shows the GRAD-CAM visualization for age-classification.  GRAD-CAM visualization suggested \textbf{different salient regions} were used for age-group classification among young, middle-aged, and older adults. \emph{Lower periocular region was used for the young-adults, upper eyelids for Middle-aged, and wrinkles around the upper eyelid and corner of the eyes for Older Adults in age-classification by ResNet-50}. A similar observation was noted for other CNN models as well. Existing studies suggest instability of iris texture over time~\cite{6516567}. Therefore, the use of the periocular region and wrinkles in age-classification from the ocular region as observed in our study supports this fact. 

\begin{table*}[]
\caption{\label{tab:young}Exact and 1-off accuracies of Age-group Classification for Young Adults.}
\centering
\resizebox{0.60\textwidth}{!}{%
\begin{tabular}{|l|l|l|l|l|}
\hline
\multicolumn{1}{|c|}{\multirow{2}{*}{\textbf{CNN}}} & \multicolumn{2}{c|}{\textbf{Left Ocular}}                          & \multicolumn{2}{c|}{\textbf{Right Ocular}}                         \\ \cline{2-5} 
\multicolumn{1}{|c|}{}                     & \multicolumn{1}{c|}{\textbf{Exact [\%]}} & \multicolumn{1}{c|}{\textbf{1-off [\%]}} & \multicolumn{1}{c|}{\textbf{Exact [\%]}} & \multicolumn{1}{c|}{\textbf{1-off [\%]}} \\ \hline
\textbf{ResNet-50}                                 & 46.72              & \textbf{93}                   & 52.87             & 93.16               \\ \hline
\textbf{MobileNet-V2}                              & \textbf{54.14}              & 91.78             & 
\textbf{59.76}              & \textbf{94.9}               \\ \hline
\textbf{ShuffleNet-V2-50}                           & 52.47              & 91.16               & 45.16            & 85.27                \\ \hline
\textbf{EfficientNet-B0}                            & 28.225                             & 53.195                           & 31.6                             & 62.64                             \\ \hline
\end{tabular}
}%
\end{table*}
\begin{table*}[]
\caption{\label{tab:middle}Exact  and 1-off accuracies of Age-group Classification for Middle-Aged Adults.}
\centering
\resizebox{0.60\textwidth}{!}{%
\begin{tabular}{|l|l|l|l|l|}
\hline
\multicolumn{1}{|c|}{\multirow{2}{*}{\textbf{CNN}}} & \multicolumn{2}{c|}{\textbf{Left Ocular}}                          & \multicolumn{2}{c|}{\textbf{Right Ocular}}                         \\ \cline{2-5} 
\multicolumn{1}{|c|}{}                     & \multicolumn{1}{c|}{\textbf{Exact [\%]}} & \multicolumn{1}{c|}{\textbf{1-off [\%]}} & \multicolumn{1}{c|}{\textbf{Exact [\%]}} & \multicolumn{1}{c|}{\textbf{1-off [\%]}} \\ \hline
\textbf{ResNet-50}                                  & \textbf{31.61}               & 81.73             & \textbf{35.03}              & \textbf{78.43}              \\ \hline
\textbf{MobileNet-V2}                             & 27.95              & \textbf{86.35}              & 31.86             & 72.69              \\ \hline
\textbf{ShuffleNet-V2-50}                           & 28.46             & 75.2                & 24.61             & 47.98              \\ \hline
\textbf{EfficientNet-B0}                            & 20.93                           & 57.86                            & 19.56                           & 55.73                             \\ \hline
\end{tabular}%
}
\end{table*}

\begin{table*}[]
\caption{\label{tab:older}Exact and 1-off accuracies of Age-group Classification for Older Adults.}
\centering
\resizebox{0.60\linewidth}{!}{%
\begin{tabular}{|l|l|l|l|l|}
\hline
\multicolumn{1}{|c|}{\multirow{2}{*}{\textbf{CNN}}} & \multicolumn{2}{c|}{\textbf{Left Ocular}}                          & \multicolumn{2}{c|}{\textbf{Right Ocular}}                         \\ \cline{2-5} 
\multicolumn{1}{|c|}{}                     & \multicolumn{1}{c|}{\textbf{Exact [\%]}} & \multicolumn{1}{c|}{\textbf{1-off [\%]}} & \multicolumn{1}{c|}{\textbf{Exact [\%]}} & \multicolumn{1}{c|}{\textbf{1-off [\%]}} \\ \hline
\textbf{ResNet-50}                                  & 28                  & 82.3                & 29.485             & 71.43               \\ \hline
\textbf{MobileNet-V2}                              & \textbf{38.07}             & 79.82               & \textbf{32.93}               & \textbf{76.65}              \\ \hline
\textbf{ShuffleNet-V2-50}                           & 36.04               & \textbf{89.65}               & 18.54              & 56.85              \\ \hline
\textbf{EfficientNet-B0}                            & 4.97                            & 32.3                            & 4.74 
& 27                             \\ \hline
\end{tabular}%
}
\end{table*}

\section{Key findings}
The key findings from all the experiments are summarized as follows:
\begin{itemize}
\item Younger adults obtained \emph{performance identical} to middle-aged individuals in user verification. Older adults' performance differs slightly in terms of EER with only a $1\%$ decrease, but the performance dropped at lower FMR points. The possible reason could be due to likely inferior quality of image capture, and relatively \emph{higher inter-class similarity due to wrinkles and folds} on the skin.

\item Almost \emph{equivalent performance was noted for gender classification} among age-groups. Middle Aged adults slightly outperformed the other two groups in gender classification by about $1\%-2\%$. A possible explanation for this observation could be the stable and distinct gender cues for middle aged adults when compared to young and older adults. Further, the periocular region contains the gender cues (also confirmed in~\cite{kuehlkamp2018predicting}) which differ across males and females. 
    
    \item Younger adult population performed the best in age classification by about 25\% accuracy gap over other age-groups. This could be due to \emph{distinct variation in the features} attributed to the growing stage of the youth population over middle-aged and older adults.
    The \emph{upper and lower eyelids and wrinkles} in the periocular region play an important role in age classification across age-groups.
    \item Lastly, \emph{inference time} which is feature extraction time in millisecond (ms) was computed for all the models on three real smartphones: iPhone 6, iPhone XR, and iPhone XR. The ShuffleNet-V2-50 model which was 98x faster than ResNet-50 (the slowest model due to being 90x larger) obtained an average inference time of $1,387$ ms. MobileNet-V2 and EfficientNet-B0, which are 68x larger than ShuffleNet-V2-50, obtained an average inference time of $497$ ms.
\end{itemize}

\section{Conclusion and Future Work}
In this study, we evaluated the fairness of RGB ocular biometrics in user authentication and soft-biometric classification among age-groups. Experimental investigations on the balanced subset of the aligned version of the UFPR ocular dataset suggest the equivalent performance of ocular biometrics in user verification and gender-classification among age-groups in terms of EER and classification accuracy. 
However, the performance of older adults dropped at lower FMR points in user verification and the young population outperformed other groups in age-classification. This could be due to the inherent characteristics of inferior quality data for older adults and distinct age clues for young adults attributed to the growing stage. The overall equivalent performance for user verification and gender classification suggests the feasibility of ocular modality towards fair and trustworthy biometrics technology. However, more experimental evaluations are required to draw any definite conclusions.
As a part of future work, further experimental validations will be performed on different ocular biometric datasets acquired across the spectrum. Comparative evaluation will be performed with face biometrics.

\section{Acknowledgement}
This work is supported from a National Science Foundation (NSF) SaTC Award \#2129173 on Probing Fairness of Ocular Biometrics Methods Across Demographic Variations. 

\small{
\bibliographystyle{plain}
\bibliography{biblio, mybibliography}
}

\end{document}